\documentclass[runningheads]{llncs}

\usepackage{graphicx}

\usepackage[utf8]{inputenc}
\usepackage{multirow}
\usepackage{booktabs}
\usepackage{array}
\usepackage{tabularx}
\usepackage{tikz}
\usepackage{amssymb}
\usepackage{color}
\usepackage{gensymb}
\usepackage{float}
\usepackage{cite}

\usepackage{graphicx}
\usepackage{subcaption}
\begin{document}
\title{Data Augmentation for Skin Lesion Analysis}
%
%
\author{F\'abio Perez$^1$ \and Cristina Vasconcelos$^2$ \and Sandra Avila$^3$ \and Eduardo Valle$^1$}
\institute{
$^1$RECOD Lab, DCA, FEEC, University of Campinas (Unicamp), Brazil\\
$^2$Computer Science Department, IC, Federal Fluminense University (UFF), Brazil \\
$^3$RECOD Lab, IC, University of Campinas (Unicamp), Brazil\\}

\authorrunning{F. Perez et al.}

\maketitle              

\begin{abstract}
Deep learning models show remarkable results in automated skin lesion analysis. However, these models demand considerable amounts of data, while the availability of annotated skin lesion images is often limited. Data augmentation can expand the training dataset by transforming input images. In this work, we investigate the impact of 13~data augmentation scenarios for melanoma classification trained on three CNNs (Inception-v4, ResNet, and DenseNet). Scenarios include traditional color and geometric transforms, and more unusual augmentations such as elastic transforms, random erasing and a novel augmentation that mixes different lesions.
We also explore the use of data augmentation at test-time and the impact of data augmentation on various dataset sizes. 
Our results confirm the importance of data augmentation in both training and testing and show that it can lead to more performance gains than obtaining new images. The best scenario
results in an AUC of 0.882 for melanoma classification without using external data, outperforming the top-ranked submission (0.874) for the ISIC Challenge 2017, which was trained with additional~data.

\end{abstract}

\section{Introduction}

\vspace{-0.02cm}
Deep learning has achieved impressive results in computer vision tasks, including skin lesion analysis~\cite{fornaciali2016towards}. However, deep learning models are data-hungry, and collecting and annotating skin lesion images can be challenging.

In image classification tasks, knowledge transfer and data augmentation are regularly employed for small datasets. Knowledge transfer usually takes place by initially training a Convolutional Neural Network (CNN) in a large source dataset (e.g., ImageNet) and using its weights as a starting point for training in the smaller target dataset~\cite{menegola2017knowledge}. Data augmentation goal is to add new data points to the input space by modifying training images while preserving semantic information and target labels. Thus, it is used to reduce overfitting.

In this work, we: (i) investigate the impact of applying diverse data augmentation techniques to three different CNN architectures (namely Inception-v4~\cite{szegedy2017inception}, ResNet~\cite{he2016deep}, and DenseNet~\cite{huang2017densely}); (ii) investigate the impact of data augmentation on different dataset sizes; and (iii) evaluate the use of different data augmentation methods during test-time, aiming to reduce generalization error. 
We conducted the experiments on the ISIC Challenge 2017 dataset~\cite{DBLP:journals/corr/abs-1710-05006} for melanoma classification~task.

\section{Related Work}

Data augmentation is broadly used in CNN architectures, such as AlexNet~\cite{DBLP:conf/nips/KrizhevskySH12}, Inception~\cite{DBLP:conf/cvpr/SzegedyLJSRAEVR15,DBLP:conf/icml/IoffeS15,szegedy2017inception}, ResNet~\cite{he2016deep}, and DenseNet~\cite{huang2017densely}. These architectures are trained on the ImageNet dataset
, which contains millions of annotated images. Some examples of data augmentation techniques are color modifications and geometric transforms (rotation, scaling, random cropping).

Models can also benefit from data augmentation on test-time. Krizhevsky et al.~\cite{DBLP:conf/nips/KrizhevskySH12} average the predictions on 10 patches (cropped from the center plus the four corners and then flipped) extracted from each test image. Szegedy et al.~\cite{DBLP:conf/cvpr/SzegedyLJSRAEVR15} report gains with a method that generates 144 patches by cropping images at different resolutions, when compared with the 10-crop method. These methods are commonly used in competitions to increase final performance but can be expensive for production.

Data augmentation is also extensively employed in skin lesion classification, a task that has much less available training data. Data augmentation is ubiquitous among top-ranked submissions in the ISIC Challenge 2017~\cite{DBLP:journals/corr/MenegolaTFLAV17,DBLP:journals/corr/BiKAF17,DBLP:journals/corr/MatsunagaHMK17}. 

Some works specifically explore data augmentation for skin lesion analysis~\cite{vasconcelos2017experiments,datadepthdesign, pham2018deep}. 
Vasconcelos and Vasconcelos~\cite{vasconcelos2017experiments} report gains in performance by using data augmentation with geometric transforms (rotations by multiples of 90 degrees; flips; lesion-preserving crops), PCA-based color augmentation, and specialist warping that preserves lesions symmetries and anti-symmetries.
Valle et al.~\cite{datadepthdesign} highlight the importance of using data augmentation for both training and testing. They averaged the predictions for 50 augmented test samples.
Pham et al.~\cite{pham2018deep} compare the effects of data augmentation on classifiers (SVM, neural networks, and random forest) trained with features extracted with a pretrained Inception-v4. Their results indicate that using more samples in test data augmentation (100 vs. 50) increases the model's performance.

In this work, we further investigate the use of data augmentation for skin lesion analysis, by comparing: test techniques (testing on a single image; test data augmentation; and test cropping, commonly employed in CNN architectures for image classification); 13 different data augmentation scenarios, including a novel augmentation; and the effects of data augmentation on different dataset~sizes.

\section{Methodology}

\subsection{CNN Architectures}
We evaluated every experiment on three very deep CNNs that are widely used in computer vision problems: Inception-v4~\cite{szegedy2017inception}, ResNet-152~\cite{he2016deep}, and DenseNet-161~\cite{huang2017densely}. We chose these networks as they achieve increased depth with different design choices and represent the state of the art in image classification.

The Inception-v4~\cite{szegedy2017inception} architecture has modules that concatenate feature maps from parallel convolutional blocks, leading to increased width and depth. Residual Networks (ResNets)~\cite{he2016deep} use shortcut connections between layers, allowing even deeper networks. Densely Connected Networks (DenseNets)~\cite{huang2017densely} concatenate the output of each layer to all subsequent layers inside a dense block, increasing the parameter efficiency and reducing overfitting. 

Since we used the same optimization hyperparameters for the three networks, we do not intend to compare the numeric values alone, but rather compare big-picture results and trends.

\subsection{Data Augmentation Techniques}
\begin{figure}[ht]
  \fontsize{30}{30}\selectfont
\centering
  \includegraphics[width=\textwidth]{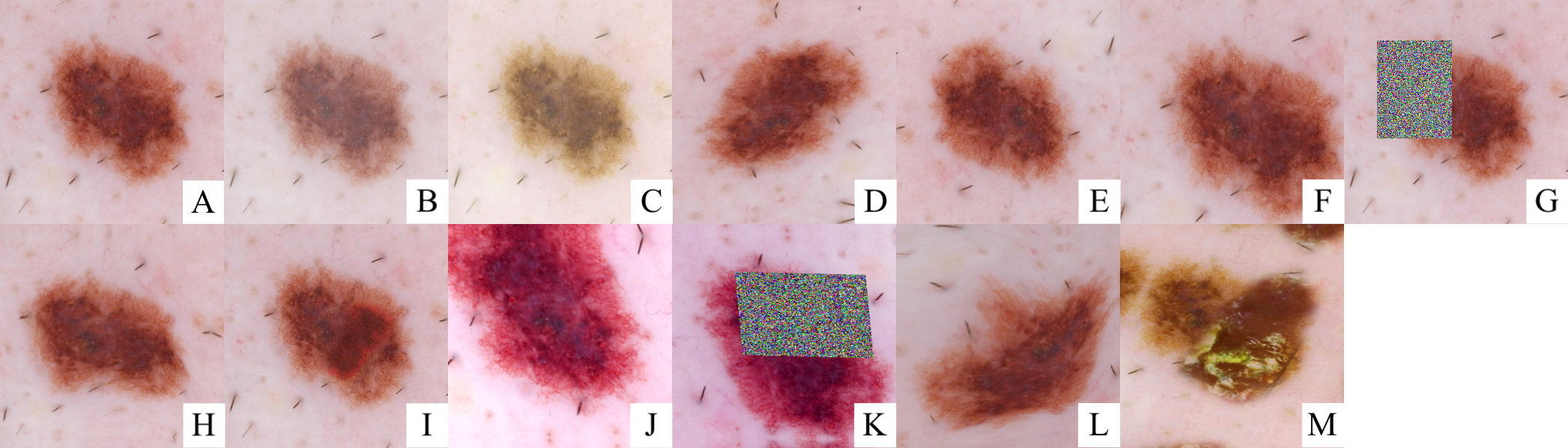}
  \caption{Examples of augmentation scenarios, described in Table~\ref{table:augmentations-techniques}.}
  \label{fig:augs}
\end{figure}

We evaluated 13 data augmentation scenarios, comprising different image processing techniques, and some combinations of them. Table~\ref{table:augmentations-techniques} describes the implementation details for each scenario. Fig.~\ref{fig:augs} shows examples of all scenarios.

\newcommand*{\MOTIVATION}{} 
\begin{table}[h!]
\centering
\caption{Augmentation scenarios. Scenarios \textbf{J} to \textbf{M} represent augmentations compositions applied in the presented order.}
\label{table:augmentations-techniques}
\begin{tabular}{>{\centering\arraybackslash}p{0.05\linewidth}>{\raggedright\arraybackslash}p{0.26\linewidth}p{0.69\linewidth}}
\textbf{ID} & \multicolumn{1}{l}{\textbf{Name}}         & \multicolumn{1}{l}{\textbf{Description}} \\ \hline
\textbf{A}  & No Augmentation & No data augmentation. Only preprocess images, as described in Section~\ref{section:training}. \\ \hline
\textbf{B}  & Saturation, Contrast, and Brightness      & Modify saturation, contrast, and brightness by random factors sampled from an uniform distribution of $[0.7, 1.3]$\ifdefined\MOTIVATION, simulating changes in color due to camera settings and lesion characteristics.\else.\fi \\ \hline
\textbf{C}  & Saturation, Contrast, Brightness, and Hue & As described in B, but also shift the hue by a value sampled from an uniform distribution of $[-0.1, 0.1]$. \\ \hline
\textbf{D}  & Affine & Rotate the image by up to $90^{\circ}$, shear by up to $20^{\circ}$, and scale the area by $[0.8, 1.2]$. New pixels are filled symmetrically at edges. \ifdefined\MOTIVATION This can reproduce camera distortions and create new lesion shapes.\fi \\ \hline
\textbf{E}  & Flips & Randomly flip the images horizontally and/or vertically. \\ \hline
\textbf{F}  & Random Crops & Randomly crop the original image. The crop has $0.4-1.0$ of the original area, and $3/4 - 4/3$ of the original aspect ratio. \\ \hline
\textbf{G}  & Random Erasing & Fill part of the image (area up to 30\% of the original image) with random noise. The transformation is applied with a probability of 0.5. Implemented as described in~\cite{randomerasing}. \ifdefined\MOTIVATION The network may benefit from occlusion by learning to look for different lesion attributes.\fi \\ \hline
\textbf{H}  & Elastic & Warp images with Thin Plate Splines (TPS). The warp is generated by defining the origins as an evenly-spaced $4\times4$ grid of points, and destinations as random points around the origins (by up to $10\%$ of the image width on each direction). \ifdefined\MOTIVATION This can produce new lesion shapes while maintaining medical attributes.
\\ \hline
\textbf{I}  & 
Lesion Mix & Mix two lesions, by inserting part of a foreground lesion (cut by its segmentation mask) into a background lesion. We apply Gaussian blur to the foreground lesion to avoid sharp edges, and equalize its color histogram with respect to the segmented background lesion. The resulting image is labeled as melanoma only if one of the two original lesions was labeled as melanoma. \ifdefined\MOTIVATION This can simulate clinical conditions with two lesions occur at the same location. \fi We did not apply this transform at test-time.\\ \hline
\textbf{J}  & 
Basic Set &  
F $\rightarrow$ D $\rightarrow$ E $\rightarrow$ C. \\ \hline
\textbf{K}  & 
Basic Set + Erasing & 
F $\rightarrow$ G $\rightarrow$ D $\rightarrow$ E $\rightarrow$ C. \\ \hline
\textbf{L}  &
Basic Set + Elastic & 
F $\rightarrow$ D $\rightarrow$ H $\rightarrow$ E $\rightarrow$ C.                            \\ \hline
\textbf{M}  & 
Basic Set + Mix 
& I $\rightarrow$ F $\rightarrow$ D $\rightarrow$ E $\rightarrow$ C. \\ \hline
\end{tabular}
\end{table}
\vspace{-0.35cm}

\subsection{Training and Evaluation}
\label{section:training}
We trained each network with Stochastic Gradient Descent (SGD) with a momentum factor 0.9, batch size of 32, starting learning rate 1e-3, reduced to 1e-4 after the $10^{th}$ epoch. The training data was shuffled before each epoch. The networks were initialized with weights trained on the ImageNet dataset, and fine-tuned with the ISIC Challenge 2017 train dataset (2000 images) \cite{DBLP:journals/corr/abs-1710-05006}. The experiments were implemented with PyTorch (\url{pytorch.org}). Augmentations were implemented with torchvision and imgaug (\url{github.com/aleju/imgaug}).

All images were resized offline to a maximum width or height of 1024 pixels to avoid expensive resizing during training. On training, images were resized to the default input sizes for each network ($224\times224$ for DenseNet and ResNet; $299\times299$ for Inception-v4), although larger sizes were possible due to global average pooling. 
Images were normalized (subtract from the mean and divide by the standard deviation) based on the ImageNet dataset, in which the networks were pretrained. Augmentations were randomly applied online during training.

We applied early stopping to interrupt the training, monitoring the AUC value for the ISIC Challenge 2017 official validation dataset (150 images) for each epoch. 
The AUC value was calculated by averaging the predictions for 16 randomly augmented copies of each validation image, by applying the same transforms used during training. The early stopping monitor interrupted the training when the validation AUC did not improve after 8 epochs. The final test AUC was calculated on the ISIC Challenge 2017 official test dataset (600 images) in three different ways: i) inputting the original test images to the network; ii) averaging the predictions for 64 randomly augmented copies of each test image; iii) averaging the predictions for 144 patches produced by cropping each test image as described in \cite{DBLP:conf/cvpr/SzegedyLJSRAEVR15}. The weights used for testing were selected from the best AUC in the validation dataset. The validation-time and test-time augmentations followed the same transforms as the training.

For every setup, we run 6 separate trainings to reduce the effects of randomness. We used Sacred (\url{github.com/IDSIA/sacred}) to organize all experiments. 

To guarantee reproducibility, we provide the documented source code used in the experiments (\url{github.com/fabioperez/skin-data-augmentation}).\vspace{-0.2cm}

\section{Results and Discussion}

\subsection{Augmentation on Training and Testing}

In this section, we discuss the results of train and test data augmentation for the proposed scenarios. Fig.~\ref{fig:aug_results} summarizes the results.

\definecolor{color:triangle}{HTML}{008cff}
\definecolor{color:square}{HTML}{ff00a1}
\definecolor{color:circle}{HTML}{777777}
\begin{figure}[bh]
  \fontsize{30}{30}\selectfont
  \begin{center}
      \includegraphics[width=\textwidth]{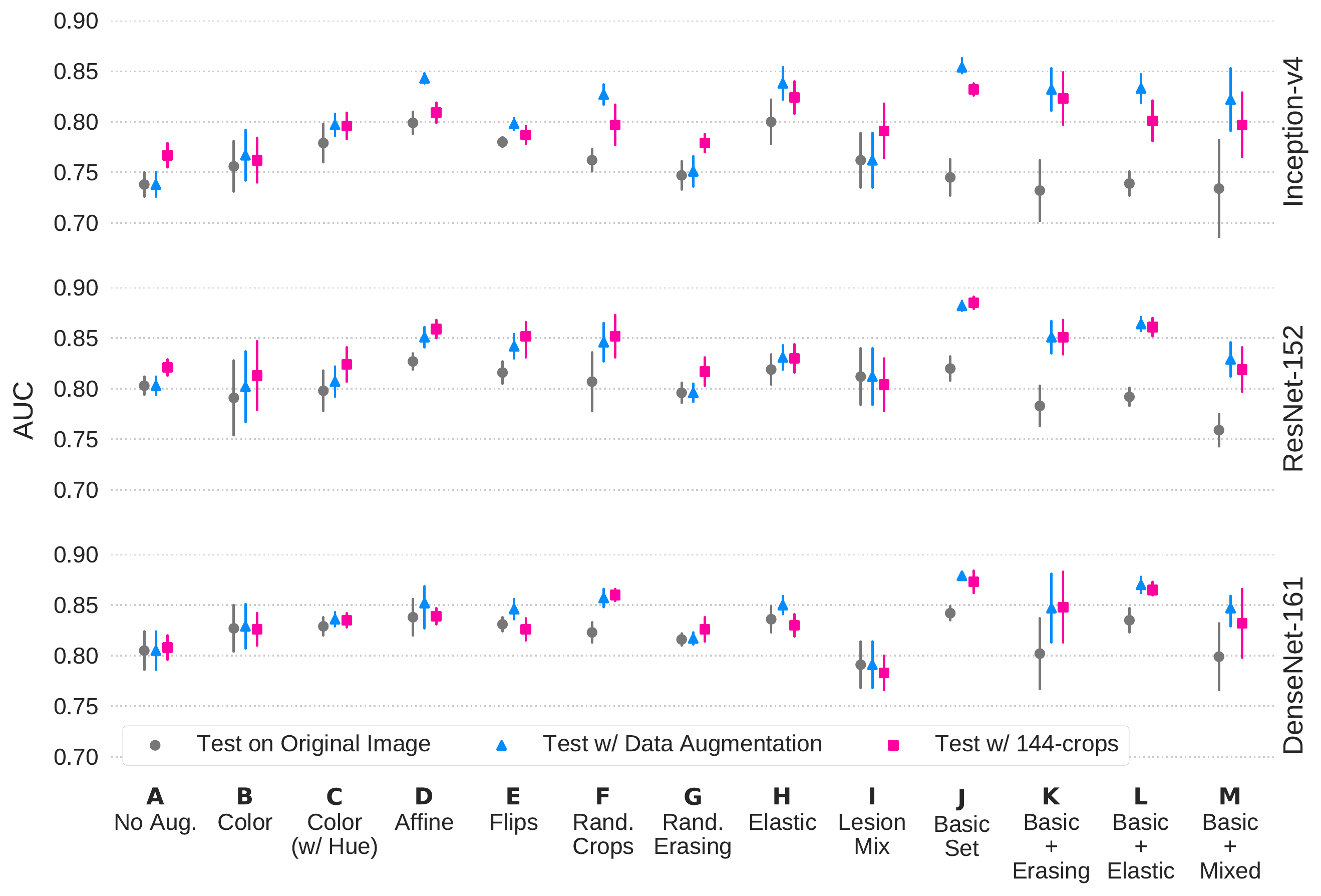}      
  \end{center}
  \vspace{-0.4cm}
  \caption{Mean AUC values for augmentation scenarios. Each color and marker represent a prediction method: \textcolor{color:circle}{$\bullet$} original image; \textcolor{color:triangle}{$\blacktriangle$} test-time data augmentation (64 images); \textcolor{color:square}{$\blacksquare$} 144 crops. Error bars represent the standard deviation for 6 runs. Values reported on ISIC Challenge 2017 test set.\vspace{-0.2cm}}
  \label{fig:aug_results}
\end{figure}

Scenario C (saturation, contrast, brightness, and hue) resulted in better AUC than scenario B (saturation, contrast, brightness) for all three networks. However, both color transforms performed worse than scenario A (no augmentation) with 144 crops on ResNet. Geometric transforms --- affine (B), random crops (F), and elastic transformations (H) --- had more consistent improvements among all three networks.

Random erasing (G) shows little improvements for Inception and DenseNet, but produce worse results than scenario A (no augmentation) with ResNet. Using 144 crops was better than test data augmentation, probably due to the destructive behavior of the method. When combined with other transformations (scenario K), random erasing reduced the test AUC in comparison with scenario~J (basic set combining traditional augmentations).

Scenario~H (elastic) shows promising results, but when applied with other common augmentation techniques (L) also performed worse than scenario~J. This may occur due to deformations produced by the combined augmentation.

Lesion mix (I~and~M) had worse performances when compared to other augmentations, indicating that the generated images were not useful. We presume that the produced images were not able to preserve relevant features from both source lesions.

Scenario J (basic set) yields the best AUC values for all three networks: 0.854 for Inception-v4, 0.882 for ResNet, and 0.879 for DenseNet. The top-ranked submissions for melanoma classification scored 0.874~\cite{DBLP:journals/corr/MenegolaTFLAV17}, 0.870~\cite{DBLP:journals/corr/BiKAF17}, 0.868~\cite{DBLP:journals/corr/MatsunagaHMK17}. They used, respectively, $9640$, $~9600$, and $3444$ images for training. Our method achieved a higher AUC with ResNet and DenseNet without additional data. Scenario J also has the highest AUC for the validation set in all three networks.

For every scenario, averaging augmented samples or 144 crops resulted in better performance than predicting on the original image alone. Even when no data augmentation was employed during training, 144 crops significantly increased the AUC, indicating that the model can benefit from different representations of the input image.

For ResNet and DenseNet, 144 crops has similar results to using data augmentation on test-time. Considering that we used 64 augmented samples vs 144 crops, test data augmentation can lead to faster inference.

Particularly, Inception-v4 has a worse performance with 144 crops than with test data augmentation in most scenarios. This may indicate that Inception-v4 suffers from overfitting, considering that data augmentation produced similar patterns on both training and testing.

\subsection{Impact of Data Augmentation on Different Dataset Sizes}

We trained each network on random subsets of 1500, 1000, 500, 250, and 125 images of the original data to analyze the effects of having limited training data. We generated a random subset for each one of the 6~runs. Fig.~\ref{fig:limit_data} summarizes the~results.

\definecolor{color:triangle2}{HTML}{ff8801}
\definecolor{color:square2}{HTML}{999999}
\definecolor{color:circle2}{HTML}{008cff}
\begin{figure}[t]
  \fontsize{30}{30}\selectfont
  \begin{center}
      \includegraphics[width=\textwidth]{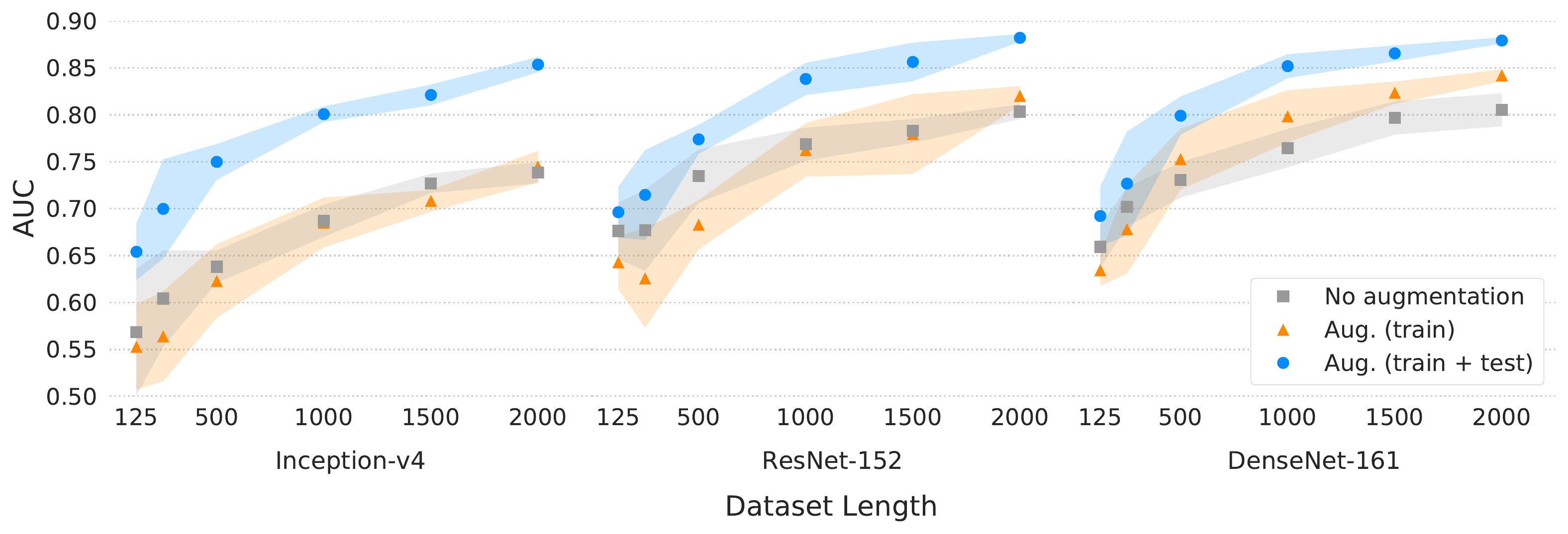}
  \end{center}
  \vspace{-0.4cm}
  \caption{Mean AUC values for different training dataset sizes, randomly sampled from the ISIC Challenge 2017 training dataset. Colors and markers represent the use of data augmentation: \textcolor{color:square2}{$\blacksquare$} no data augmentation; \textcolor{color:triangle2}{$\blacktriangle$} train data augmentation (scenario J); \textcolor{color:circle2}{$\bullet$} train and test data augmentation (scenario J, averaging each test image on 64 augmented samples). Bands represent the standard deviation for 6~runs. Values reported on ISIC Challenge 2017 test set.\vspace{-0.2cm}} \label{fig:limit_data}
\end{figure}

Applying data augmentation (scenario J) during both training and testing noticeably improved performance for datasets with 500 or more images. Data augmentation for training only worsened the results for very small data sizes ($< 500$ images) and led to little or no improvement for other sizes, showing the importance of applying data augmentation during test-time.

The impact of data augmentation on Inception-v4 was more perceptible than on other networks, which may be caused by the regularizing properties of ResNet and DenseNet architectures. Training Inception-v4 with 500 images and data augmentation resulted in better performance than training with 1000, 1500 or 2000 images without augmentation. ResNet and DenseNet achieved a higher AUC with 1000 images and data augmentation than with 1500 and 2000 images without augmentation.
This indicates that, in some cases, using data augmentation can be more effective than adding new training data. Nevertheless, employing data augmentation does not reduce the importance of adding new data, giving that the network can benefit from both.\vspace{-0.1cm}

\section{Conclusion}

The results highlight the positive impact of using data augmentation for training melanoma classification models. Moreover, models can also benefit from test data augmentation.

The best augmentation scenario (J), which combines geometric and color transformations, surpasses the top-ranked AUC values for the ISIC Challenge 2017 without any additional data. Fine-tuning hyperparameters and model ensembling may result in additional performance gains.

Lesion mix augmentation (scenarios I and M) have inferior results when compared with other scenarios. We implemented this augmentation through handcrafted image processing techniques, which may not be appropriate for producing reliable images. More advanced approaches, such as Generative Adversarial Networks or other generative architectures~\cite{bissoto2018skin}, might lead to better results.

\section*{Acknowledgments}
We gratefully acknowledge NVIDIA Corporation for the donation of GPUs and Microsoft Azure for the GPU-powered cloud platform used in this work. C. Vasconcelos and E. Valle are partially funded by Google Research LATAM 2017. E. Valle is also partially funded by CNPq PQ-2 grant (311905/2017-0) and Universal grant (424958/2016-3). RECOD Lab. is partially supported by diverse projects and grants from FAPESP, CNPq, and CAPES.

%
%
%
\bibliographystyle{splncs04}
\bibliography{references}
\end{document}